 \documentclass[sigconf, 10 pt]{acmart}
\renewcommand\footnotetextcopyrightpermission[1]{} 
\usepackage{graphicx}
\usepackage{amsmath}
\usepackage{mathtools}
\usepackage{comment}
\usepackage{titlesec}

\titleformat{\subsubsection}
  {\normalfont\normalsize\bfseries}
  {\thesubsubsection}
  {1em}
  {}

\setcitestyle{square, comma, numbers,sort&compress, super}

\usepackage[subrefformat=parens,labelformat=parens]{subfig}
\usepackage{bm}
\usepackage{multirow}
\usepackage{threeparttable,booktabs}
\usepackage{tikz}
\usepackage{balance}
\usepackage{courier}                                       
\usepackage{cleveref}                                      
\usepackage[mathcal]{eucal}
\usepackage[noend]{algpseudocode}
\algrenewcommand\textproc{\texttt}
\makeatletter\let\float@addtolists\relax\makeatother
\usepackage{algorithm}
\usepackage{pgfplots}
\usepackage{pgfplotstable}
\pgfplotsset{compat=newest}
\usepackage{siunitx}
\theoremstyle{plain}

\theoremstyle{definition}

\usepackage[skip=1pt]{caption}            
\setcitestyle{square, comma, numbers, sort&compress, super}

 \graphicspath{{./figs/}}

\begin{document}

\title{\textbf{INSIGHT:} 
Un\underline{i}versal \underline{N}eural \underline{Si}mulator for Analo\underline{g} Circuits \underline{H}arnessing Autoregressive
\underline{T}ransformers
}



\author{Souradip Poddar\textsuperscript{1}, Youngmin Oh\textsuperscript{2}, Yao Lai\textsuperscript{1}, Hanqing Zhu\textsuperscript{1}, \\Bosun Hwang
\textsuperscript{2}, David Z. Pan\textsuperscript{1}\\
\large{
\textsuperscript{1}ECE Department, The University of Texas at Austin, Austin, TX, USA,\\
\textsuperscript{2}Samsung Advanced Institute of Technology, Republic of Korea}\\
\normalsize{
souradippddr1@utexas.edu, youngmin0.oh@samsung.com, yao.lai@austin.utexas.edu, \\hqzhu@utexas.edu, bosun.hwang@samsung.com,
dpan@ece.utexas.edu
}
}

\begin{abstract}
Analog front-end design heavily relies on specialized human expertise and costly trial-and-error simulations, which motivated many prior works on analog design automation. 
However, \textit{efficient} and \textit{effective} exploration of the vast and complex design space remains constrained by the time-consuming nature of SPICE simulations, making effective design automation a challenging endeavor.
In this paper, we introduce INSIGHT, a GPU-powered, technology-agnostic, effective universal neural simulator in the analog front-end design automation loop.
INSIGHT accurately predicts the performance metrics of analog circuits across various technologies with just a few microseconds of inference time.
Notably, its autoregressive capabilities enable INSIGHT to accurately predict simulation-costly critical transient specifications leveraging less expensive performance metric information.
The low cost and high fidelity feature make INSIGHT a good substitute for standard simulators in analog front-end optimization frameworks. 
INSIGHT is compatible with any optimization framework, facilitating enhanced design space exploration for sample efficiency through sophisticated offline learning and adaptation techniques.
Our experiments demonstrate that INSIGHT-M, a model-based batch reinforcement learning sizing framework with INSIGHT as the accurate surrogate, only requires $<$ 20 real-time simulations with 100-1000x lower simulation costs and significant speedup over existing sizing methods.
\end{abstract}
\maketitle
\section{Introduction} \label{sec:Introduction}
\sloppy
Analog circuits vary significantly across and within types, each characterized by distinct performance metrics, making analog front-end design challenging. 
This complexity necessitates a labor-intensive process reliant on designer expertise and costly trial-and-error simulations. 
The consequent prolonged design cycles and increased time to market have motivated extensive research towards its automation.

Various knowledge-based and optimization-based approaches, including equation-based and simulation-based methods, have been proposed for analog design automation.
While knowledge-based methods, such as ~\cite{horta2002analogue,EDTC95:Veselinovic:Topology_Sel_EDTC,ICCAD08:McConaghy:Topology_Sel_ICCAD}, are time-consuming and lack scalability, equation-based methods, like ~\cite{daems2003simulation,boyd2001optimal,wang2014enabling,TCAD20:Zhao:Topology_Synthesis_TCAD}, are also highly inaccurate on modern technologies.
Conversely, simulation-based methods, while effective, depend on costly real-time simulations.
Among these, evolutionary algorithms like particle swarm optimization  ~\cite{vural2012analog}, advanced differential evolution ~\cite{liu2014automated}, and genetic algorithms ~\cite{Springer:GenAlgo1, kumar2012optimized} require extensive SPICE simulations (sample inefficient).
Deep Reinforcement Learning (RL) methods like ~\cite{choi2023reinforcement, zhang2023automated, li2021circuit,DAC20:Wang:GCNRL, DATE20:Settaluri:Autockt} while robust, also involve frequent simulator interactions.
Meanwhile, Bayesian Optimization ~\cite{wu2019hyperparameter,Snoek:snoek2012practical} suffers from scalability issues due to computational demands associated with model updates.
Modern algorithms ~\cite{DAC21:Budak:DNN_Opt, ASPDAC23:Budak:APOSTLE, DATE24:CRONus:Oh} use black-box optimization and model-based RL to enhance sample efficiency yet require numerous costly real-time simulations.
Given the vast and complex design space and intricate trade-offs among performance metrics, repeated compute-intensive simulations within the optimization loops become inherent for effective and efficient design space exploration and exploitation to ensure optimal circuit performance.

This heavy reliance on expensive real-time simulations motivates us to find an \textit{efficient} and \textit{effective} surrogate for the standard SPICE simulator to expedite analog design optimization tasks.
The complexity of parameter-performance relationships in analog circuits renders simple DNNs inadequate, necessitating bigger, more complex models and extensive, costly EDA data while still resulting in suboptimal accuracies~\cite{DATE24:CRONus:Oh}.
Given the high interdependence of performance metrics inherent to analog circuits, we propose to formulate the analog circuit performance prediction problem as an autoregressive generation task, i.e., predicting each remaining simulation-costly performance specification by leveraging provided design parameters and partial performance information (provided or previously-generated) in a sequential manner.
The autoregressive problem setup enforces the model to automatically reason and capture the inherent interdependency in the challenging analog performance prediction task, easing the model learning complexity, therefore, providing much better prediction quality than vanilla DNN setup~\cite{DATE24:CRONus:Oh}.

In this paper, we introduce INSIGHT, a novel, technology-agnostic, universal neural circuit simulator that accurately predicts performance on previously unseen parameters across various circuits and technologies autoregressively, consistently demonstrating  R\textsuperscript{2} scores as high as 0.99 and minimal test losses at microsecond inference times.
Its autoregressive capabilities facilitate accurate prediction of simulation-costly critical transient specifications leveraging performance data from less expensive simulations.
These features position INSIGHT as a cost-effective and high-fidelity alternative to standard circuit simulators in analog front-end optimization frameworks.
INSIGHT is highly scalable and can integrate into existing EDA optimization workflows to enhance efficiency. 
To showcase the effectiveness of INSIGHT as a surrogate model, we integrate INSIGHT within a model-based batch RL framework (INSIGHT-M) for circuit sizing and observe very high sample efficiencies for different circuits, covering several circuit types, like operational amplifiers, transimpedance amplifiers, comparators, and level shifters, across various technologies.

\vspace{5pt}
The key contributions of this work include:
\begin{itemize}
\item We introduce a novel problem formulation for analog circuit performance prediction by conceptualizing it as an autoregressive sequence modeling problem, which significantly eases the learning complexity of the neural model. 

\item We build INSIGHT, a GPU-accelerated, technology-agnostic universal neural simulator based on a decoder-only Transformer architecture to solve the analog circuit performance prediction problem.
INSIGHT enables accurate prediction of simulation-costly critical transient specifications using less expensive performance metrics and 
and demonstrates its potential as a low-cost, high-fidelity surrogate model in the analog design automation loop.

\item  We have rigorously validated our INSIGHT across various circuit and performance metric types across multiple technologies.
We also demonstrate INSIGHT-M, which leverages INSIGHT as the performance predictor for the device sizing problem.
INSIGHT-M achieved exceptional sample efficiencies on various circuit types, requiring $<$ 20 real-time simulations compared to existing sizing methods, which require 100-1000x more simulations.

\item To the best of our knowledge, this marks the first \textbf{successful} application of a universal neural simulator framework leveraging autoregressive Transformers for analog design automation.
\end{itemize}
\vspace{-5pt}
\section{Preliminaries} \label{sec:prelim}
In this section, we briefly discuss some insights into analog circuit basics/metrics (Section ~\ref{Analog_Physics}) and Transformer architectures (Section ~\ref{subsec:Transformer_basics}), discussing our rationale for choosing them.
Additionally, we define a key design quality metric, the
Figure of Merit (FoM), which will guide our sizing flow (Section ~\ref{Sizer}).

\subsection{Analog Circuit Basics and Metrics}\label{Analog_Physics}
In analog circuits, performance metrics and parameters are highly interdependent. By leveraging these interdependencies, we can build effective and data-efficient prediction models.
For instance, fundamental metrics such as Quiescent Current ($I_Q$) and DC Gain, when combined with parameter information, can significantly enhance predictions for other key specifications, like unity-gain bandwidth (UGBW) and phase margin (PM).
Similarly, integrating all static metrics provides crucial insights into the circuit's potential transient behavior. Thus, leveraging less expensive performance metric information can significantly enhance prediction accuracies for critical, simulation-costly, transient specifications such as slew rate and settling time.
Furthermore, given the transferable nature of circuit physics across technologies for a given topology, understanding performance interdependencies within one node provides valuable insights applicable to other nodes.
These insights, combined with the adaptability and transferability of Transformer architectures, motivate the adoption of autoregressive Transformers to build a universal neural simulator that effectively predicts performance across technologies.
\subsection{Transformer Architectures}\label{subsec:Transformer_basics}
Transformers constitute a powerful class of generative deep learning architectures that have revolutionized sequence modeling tasks in natural language processing (NLP) and beyond.
These models stand out due to their self-attention mechanism, which effectively captures dependencies between tokens in a sequence regardless of distance.
The self-attention mechanism evaluates dependencies within a sequence by computing a weighted sum of value vectors. These weights highlight the most relevant token information for predicting subsequent tokens and are derived from the dot product of input vectors.
Unlike traditional RNNs~\cite{rumelhart1986parallel}, Transformers process data in parallel, enhancing efficiency and scalability. 
While the original Transformer architecture ~\cite{vaswani2017attention} consists of stacked layers of encoders and decoders, decoder-only models like GPT~\cite{radford2018improving} have recently demonstrated notable success in sequence data generation. 
Decoder-only Transformers employ solely the decoder component, composed of multi-head self-attention and position-wise feed-forward networks. 

\subsection{Figure of Merit}\label{Sizer}
Given an objective performance metric \( f_0(\mathbf{x}) \) and \( m \) constraint metrics \( \{ f_i(\mathbf{x}) \leq 0 \mid i = 1, \ldots, m \} \), we adopt the Figure of Merit (FoM) definition given in ~\cite{DAC21:Budak:DNN_Opt, mlapp22:Budak:mlsizing, ASPDAC23:Budak:APOSTLE}, to evaluate quality of a design. This is of the following form:
\begin{equation}
    \label{eq:scalarizationfunc}
    \text{FoM}(\mathbf{x}) = w_0 \cdot f_0(\mathbf{x}) + \sum_{i=1}^{m} \min\left(1, \max(0, w_i \cdot f_i(\mathbf{x}))\right)
\end{equation}
where \( w_i \) is the weighting factor, a \( \max(\cdot) \) function is used for equating designs after constraints are met, and \( \min(\cdot) \) is used to prevent a single constraint violation from dominating the FoM value. This formulation will guide our sizing workflow.

\section{INSIGHT Framework and Algorithms} \label{sec:algo}
\begin{figure*}[t]
    \centering 
    \includegraphics[width=0.8\textwidth]{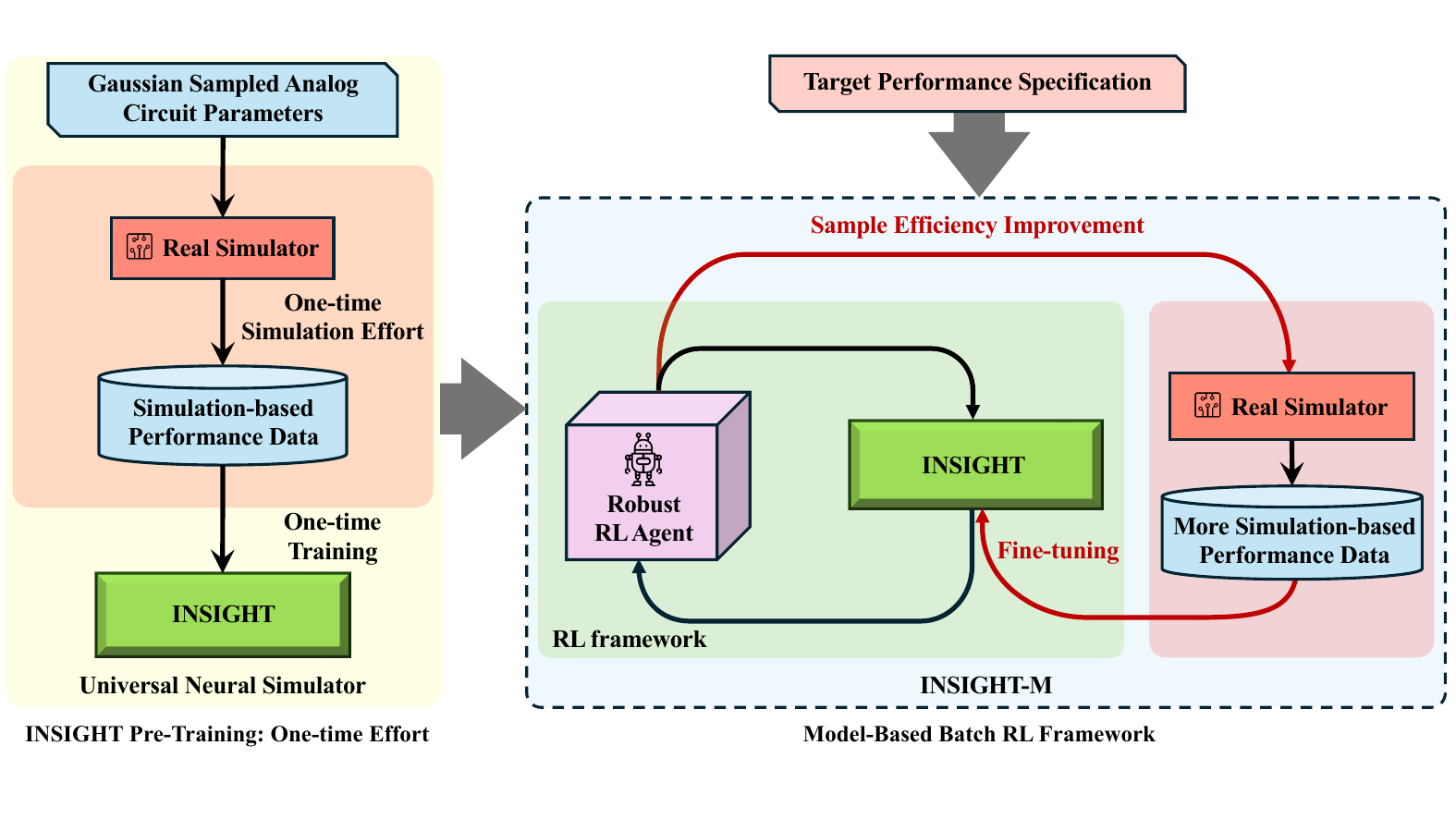}
    \vspace{-5pt}
    \caption{Automatic Analog Sizing Framework}
    \label{fig:INSIGHT-M}
\end{figure*}

In this section, we introduce our proposed frameworks.
We begin by detailing the construction of INSIGHT, our proposed universal neural simulator for analog circuits, along with the probabilistic rationale for this approach, in Section~\ref{subsec:INSIGHT}.
Section ~\ref{subsec:optimization} then details our analog circuit sizing optimization framework for a given topology (Figure~\ref{fig:INSIGHT-M}).

\subsection{INSIGHT for Modeling}\label{subsec:INSIGHT}
In INSIGHT, we are driven by a critical inquiry: Can we develop an \textit{effective} and \textit{efficient} surrogate model that utilizes circuit parameters and inexpensive partial performance metrics information (if available) to predict remaining simulation-costly performance numbers accurately? This question is particularly pertinent as current analog front-end design automation heavily depends on time-intensive simulations like SPICE, incurring a strong demand to ease the reliance.

The problem 
can be formally represented as follows:
For a given analog circuit topology, given a sequence of design parameters (\( x_1 \) to \( x_N \)), such as width-to-length ratios (W/L), our goal is to predict the corresponding performance specifications (\( y_1 \) to \( y_M \)) accurately. 
This predictive task is mathematically framed as maximizing the conditional probability $P(y_1, y_2, \ldots, y_M \mid x_1, x_2, \ldots, x_N)$, which can be decomposed into a product of conditional probabilities as follows:
\begin{equation}
\begin{split}
P(y_1, y_2, \ldots, y_M \mid x_1, x_2, \ldots, x_N) = \\
\prod_{i=1}^M P(y_i \mid x_1, x_2, \ldots, x_N, y_1, y_2, \ldots, y_{i-1})
\end{split}
\end{equation}
This objective is equivalent to minimizing the Negative Log Likelihood (NLL), expressed as:
\begin{equation}
\text{NLL} = -\sum_{i=1}^M \log P(y_i \mid x_1, x_2, \ldots, x_N, y_1, y_2, \ldots, y_{i-1})
\end{equation}

\begin{figure}[h]
    \centering 
    \includegraphics[width=0.7\columnwidth]{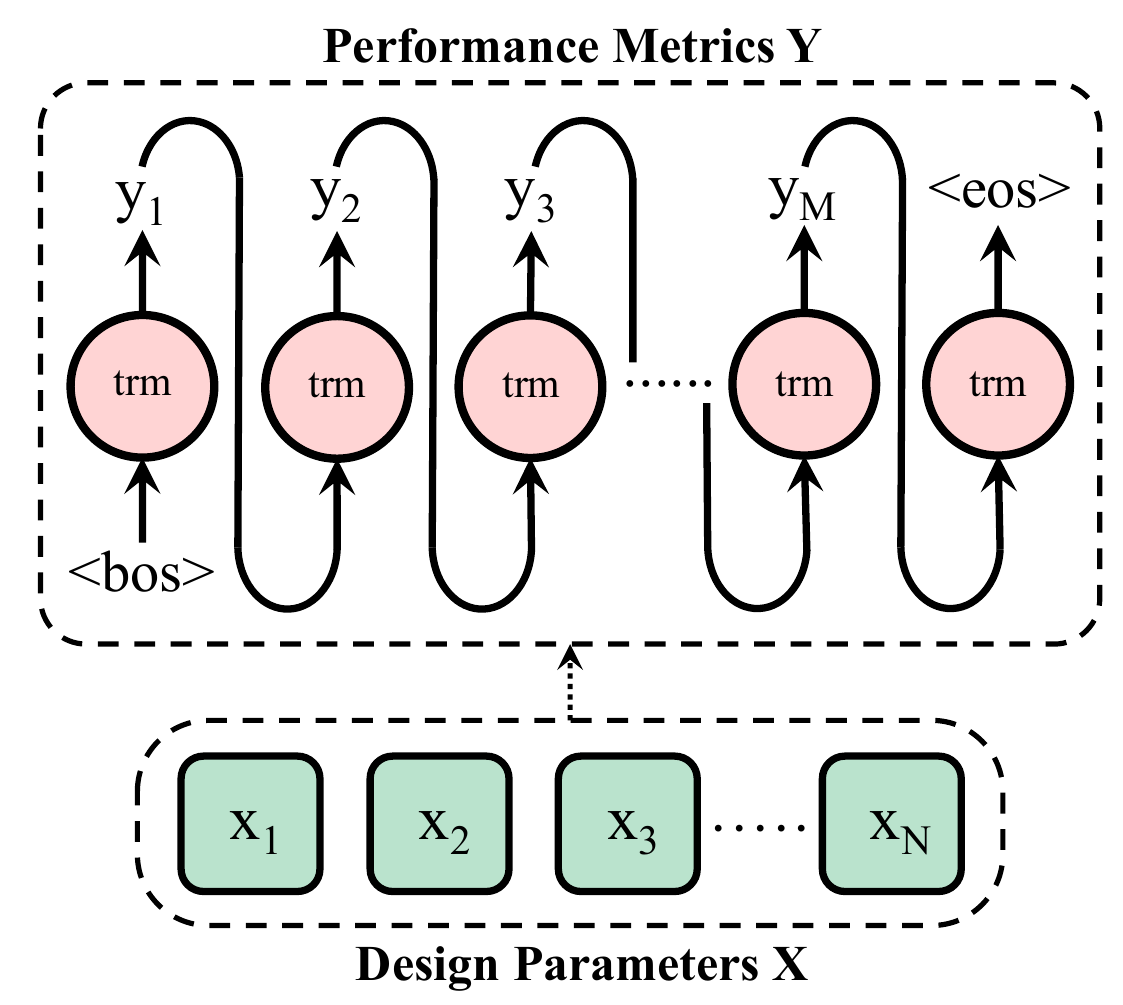}
    \vspace{5pt}
    \caption{Autoregressive Performance Prediction in INSIGHT}
    \label{fig:INSIGHT}
\end{figure}

Each term, $P(y_i | x_1, x_2, \dots, x_{N}, y_1, y_2, \ldots, y_{i-1})$, quantifies the likelihood of observing the token $y_i$, given the design parameters and the predicted performance specifications.
This formulation connects to the contextual approach in autoregressive sequence generation tasks involving decoder-based Transformers like GPT~\cite{radford2018improving} and motivates us to use a decoder-style Transformer model for INSIGHT.
Figure~\ref{fig:INSIGHT} presents a conceptual diagram illustrating INSIGHT's autoregressive sequential inferencing of performance metrics.

In autoregressive sequence generation, accurate prediction of one metric significantly enhances the prediction accuracy of the subsequent, emphasizing the importance of metric sequence order for effective and data-efficient predictions. 
We employ a greedy strategy, ordering metrics by increasing complexity and dataset needs. Specifically, we start with simpler, cost-effective metrics such as DC and AC and progress to more complex ones like transients. Also, if a specification is known to be directly derived from others, it is sequenced after the latter to optimize data need and prediction accuracy.
By avoiding the early use of complex metrics, which demand more data, this approach leverages simpler specifications to enhance the prediction accuracy of more complex ones, ensuring optimal, effective, and data-efficient sequencing. 

Given a performance metric sequence order, INSIGHT utilizes self-attention to adaptively focus on the most relevant segments of the parameter (\(x_1\) to \(x_N\)) and generated performance output (\(y_1\) to \(y_{i-1}\)) sequence, to accurately predict each output metric \(y_i\). Positional encoding is used to preserve sequence order and maintain context and coherence. 
Table~\ref{tab:hyperparam} presents INSIGHT's architecture details.

\vspace{-5pt}
\begin{table}[h!]
\begin{center}
\caption{INSIGHT Architecture}
\label{tab:hyperparam}
\begin{small}
\begin{tabular}{c|c}
\toprule
INSIGHT Parameter & Value \\
\midrule
\hspace{0.5mm} Network Architecture & Decoder-Based Transformer\\
\hspace{0.5mm} Dim. of model & 76\\
\hspace{0.5mm} \# of Heads & 4\\
\hspace{0.5mm} \# of Decoder Layers & 3\\
\hspace{0.5mm} Activation & GeLU~\cite{hendrycks2016gaussian}\\
\hspace{0.5mm} Learning rate & 0.001  \\& with cosine annealing scheduler ~\cite{sgdr}\\
\bottomrule
\end{tabular}
\end{small}
\end{center}
\vspace{-8pt}
\end{table}
\vspace{-5pt}
\begin{table}[h!]
\begin{center}
\caption{Neural Simulator Architecture in CRONuS}
\label{tab:hyperparam_fcensemble}
\begin{small}
\begin{tabular}{c|c}
\toprule
FC-Ensemble Parameter & Value \\
\midrule
\ Network Architecture & Fully-Connected\\
\# of Hidden Units & 200-200-200-200-200\\
\# of Ensemble Networks & 7\\
Activation & ReLU~\cite{relu}\\
\hspace{0.5mm} Learning rate & 0.001\\
\bottomrule
\end{tabular}
\end{small}
\end{center}
\vspace{-5pt}
\end{table}

Note: While the relationship between performance metrics, relative to design parameters and among themselves, is not inherently time-series, this formulation enables self-attention and autoregressive features to leverage their interdependencies for effective and efficient predictions. 

Although our task is analogous to sequence generation tasks, here we need to predict real values instead of generating a sequence of tokens, such as words or positions in images, from a set of finite elements or generating scalar real values from sequential contexts, as shown in previous Transformer works.
Generating a sequence of real values based on Transformer architecture is unclear ~\cite{chen2023tsmixer}, and thus, suitable foundation models for predicting circuit performance metrics do not exist. 
Therefore, evaluating the feasibility of the Transformer architecture in generating a sequence of circuit outputs is crucial to validate its potential for building domain-specific performance prediction models within a universal framework.

Building on the existing decoder-Transformer architecture~\cite{radford2018improving, radford2019language}, our proposed architecture for INSIGHT, combined with our strategy for optimizing performance sequence order, offers a data-efficient and effective universal framework that seamlessly accommodates various circuit types and topologies, is technology-agnostic and automatically adapts to diverse parameter and performance metric lengths.
By focusing on simplicity, we improve usability and scalability while maintaining robust predictions.

\subsection{INSIGHT for Analog Sizing}\label{subsec:optimization}
Figure~\ref{fig:INSIGHT-M} showcases our sizing optimization algorithm.
It comprises of two main sections: INSIGHT Pre-Training Phase
(Section ~\ref{subsubsec:Data-Collection-Training})
and INSIGHT for Optimization (Section ~\ref{subsubsec:INSIGHT-M}). 

\subsubsection{INSIGHT Pre-Training}
\label{subsubsec:Data-Collection-Training}
As depicted in the left part of Figure~\ref{fig:INSIGHT-M}, the pre-training phase involves a one-time effort to generate a simulation-based performance dataset, followed by a one-time training of our proposed neural simulator, INSIGHT.
The data collection involves gathering a random dataset of Gaussian-sampled design parameters and their corresponding performance metrics derived from SPICE simulations. 
To address computational demands, the initial dataset size is defined as a hyperparameter, allowing for adjustments tailored to specific use cases.
This flexibility helps balance the resource usage during the initial data collection with the subsequent sample efficiency following a user query. 
Additionally, INSIGHT can leverage transfer learning to reduce dataset requirements across technologies while maintaining high performance.

\begin{figure*}[!t]
\centering
\setlength{\fboxsep}{0pt} 
\setlength{\fboxrule}{0.5pt} 
\begin{tabular}{@{}c@{\hspace{-0.8pt}}c@{\hspace{-0.8pt}}c@{\hspace{-0.8pt}}c@{\hspace{-0.8pt}}c@{\hspace{-0.8pt}}c@{}} 
\fbox{\includegraphics[height=0.81in]{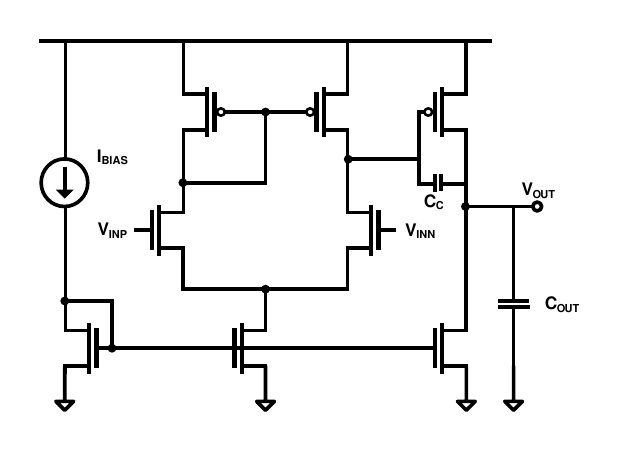}}&
\fbox{\includegraphics[height=0.81in]{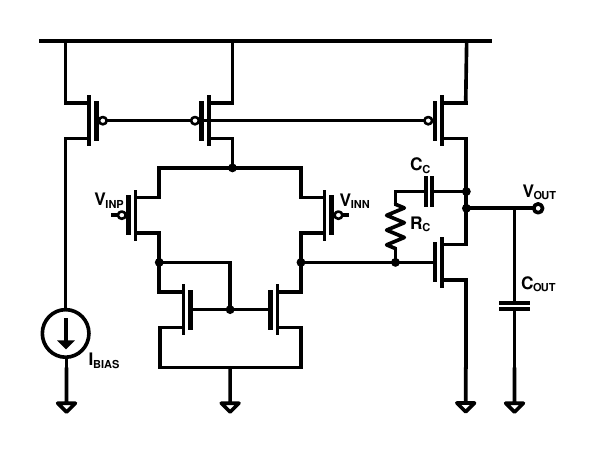}}&
\fbox{\includegraphics[height=0.81in]{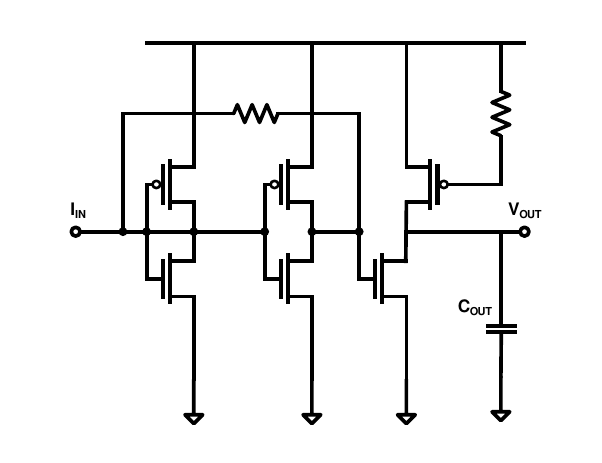}}&
\fbox{\includegraphics[height=0.81in]{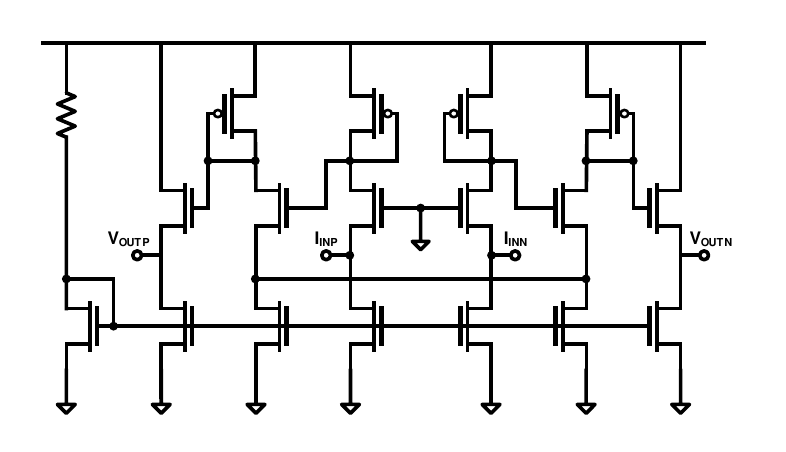}}&
\fbox{\includegraphics[height=0.81in]{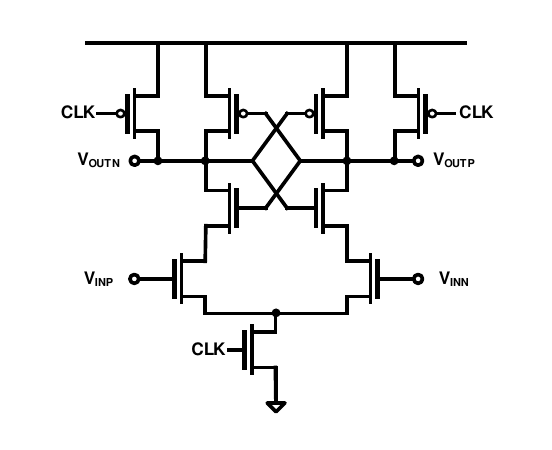}}&
\fbox{\includegraphics[height=0.81in]{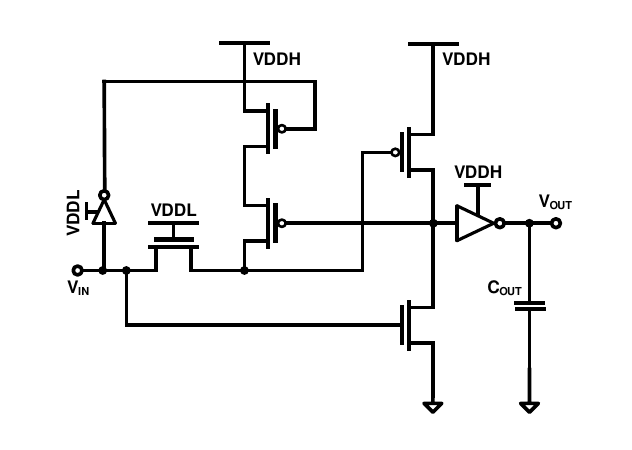}}\\
(a) & (b) & (c) & (d) & (e) & (f) \\
\end{tabular}
\caption{\small Benchmark Circuits: (a) 2-Stage Miller-Compensated OPAMP with NMOS Input Pair, (b) 
2-Stage Miller-Compensated OPAMP with PMOS Input Pair, (c) 2-Stage TIA, (d) 3-Stage TIA, (e) Comparator, (f) Level Shifter.
}
\label{fig:circuits_bench}
\end{figure*}
\vspace{5pt}
\begin{table*}[]
\footnotesize
\caption{Performance Metrics for each circuit}
\resizebox{\textwidth}{!}{
\begin{tabular}{|c|cccc|}
\hline
Circuits  & \multicolumn{4}{c|}{Performance Metrics}                                      \\ \hline
\shortstack[c]{2-Stage Miller-Compensated OPAMP with NMOS Input Pair}    & \multicolumn{1}{c|}{\multirow{4}{*}{$I_Q$ (mA)}} & \multicolumn{1}{c|}{\multirow{4}{*}{DC Gain (dB)}} & \multicolumn{1}{c|}{\multirow{4}{*}{UGBW (MHz)}} & \multirow{4}{*}{Phase Margin ($^\circ)$}\\ \cline{1-1} 
\shortstack[c]{2-Stage Miller-Compensated OPAMP with PMOS Input Pair} & \multicolumn{1}{c|}{}                            & \multicolumn{1}{c|}{}                           & \multicolumn{1}{c|}{}                              &                                                        \\ \cline{1-1} 
\shortstack[c]{2-Stage TIA}     & \multicolumn{1}{c|}{}                            & \multicolumn{1}{c|}{}                           & \multicolumn{1}{c|}{}                              &                                                        \\ \cline{1-1}
\shortstack[c]{3-Stage TIA}    & \multicolumn{1}{c|}{}                            & \multicolumn{1}{c|}{}                           & \multicolumn{1}{c|}{}                              &                                               \\ \hline
Comparator       & \multicolumn{2}{c|}{DC Power (W)}                                                                & \multicolumn{2}{c|}{Avg. Delay (sec)}                                                            \\ \hline
Level Shifter   & \multicolumn{1}{c|}{DC Power (W)}              & \multicolumn{1}{c|}{Ratio}                      & \multicolumn{1}{c|}{Avg. Delay (sec)}              & Delay Balance (sec)        \\ \hline
\end{tabular}
}
\label{tab_perf_specs}
\end{table*}

\subsubsection{INSIGHT for Optimization}\label{subsubsec:INSIGHT-M}
The pre-trained INSIGHT is subsequently integrated into a Model-based Batch RL framework for circuit sizing, referred to as INSIGHT-M. As shown in the right part of Figure~\ref{fig:INSIGHT-M}, upon query, INSIGHT-M is deployed to optimize device sizing to meet various user-specified performance targets. 
In our experiments, we use PPO-based RL agents ~\cite{schulman2017proximal} with default hyperparameters, a method extensively used in EDA because of its robustness and effectiveness ~\cite{cao2022domain,DATE20:Settaluri:Autockt}.
PPO enhances learning robustness and safety by executing multiple parallel trajectory rollouts and aggregating them to update policies under safe constraints but at the expense of poor sample efficiency.
The INSIGHT-M framework addresses this drawback by training robust RL agents within INSIGHT's synthetic environment, which closely mimics real-world conditions. 
Once trained, these agents are deployed in the real-world simulator environment using a batch RL strategy. This process involves periodic performance evaluations in the real environment. If the agents perform adequately, their deployment persists. 
If not, newly acquired real-world data combined with efficiently sampled previously collected data is utilized to fine-tune INSIGHT, and the cycle of evaluation and adjustment is repeated. This iterative refinement process optimizes the balance between safety, robustness, and efficiency.
The high fidelity of INSIGHT ensures that the agent's learning and subsequent behaviors are closely aligned with real-world design tasks, thereby minimizing trial and error and improving sample efficiency. Incorporating robust adaptation techniques, such as uncertainty-driven Upper Confidence Bound (UCB) based exploration, further enhances the agent's adaptability to real-world environments. 
Please note that other RL algorithms can be employed with INSIGHT.
\section{Experiments} \label{sec:Results}
\begin{table*}
\footnotesize
\caption{Performance Comparison Across Different Circuits: INSIGHT vs Neural Simulator in CRONuS}
\resizebox{\textwidth}{!}{
\begin{tabular}{|c|cccc|}
\hline
Circuits (45nm)  & \multicolumn{4}{c|}{R\textsuperscript{2} Scores}                                                                                       \\ \hline
Trainset:Testset Size Range & \multicolumn{2}{c|}{300:100 - 1,500:500}                                        & \multicolumn{2}{c|}{3,000:1,000 - 9,000:3,000}                  \\ \hline
Neural Simulator     & \multicolumn{1}{c|}{INSIGHT} & \multicolumn{1}{c|}{FC-Ensemble in CRONuS} & \multicolumn{1}{c|}{INSIGHT} & FC-Ensemble in CRONuS\\ \hline
Two-Stage Miller-Compensated OPAMP with NMOS Input Pair    & \multicolumn{1}{c|}{0.9901}            & \multicolumn{1}{c|}{0.6578}            & \multicolumn{1}{c|}{0.9960}            &          0.6240   \\ \hline
Two-Stage Miller-Compensated OPAMP with PMOS Input Pair & \multicolumn{1}{c|}{0.9937}            & \multicolumn{1}{c|}{0.7270}            & \multicolumn{1}{c|}{0.9973}            &         0.7872    \\ \hline
Two-Stage Transimpedance Amplifier      & \multicolumn{1}{c|}{0.9982}            & \multicolumn{1}{c|}{0.8020}            & \multicolumn{1}{c|}{0.9933}            &     0.7784        \\ \hline
Three-Stage Transimpedance Amplifier    & \multicolumn{1}{c|}{0.9997}            & \multicolumn{1}{c|}{0.9290}            & \multicolumn{1}{c|}{0.9996}            &        0.8894     \\ \hline
Comparator       & \multicolumn{1}{c|}{0.7231}            & \multicolumn{1}{c|}{0.5537}            & \multicolumn{1}{c|}{0.998}            &            0.6535 \\ \hline
Level Shifter    & \multicolumn{1}{c|}{0.9905}            & \multicolumn{1}{c|}{-0.1554}            & \multicolumn{1}{c|}{0.9265}            &       0.1252      \\ \hline
\end{tabular}}


\resizebox{\textwidth}{!}{
\begin{tabular}{|c|cccc|}
\hline
Circuits (45nm)  & \multicolumn{4}{c|}{MSE}                                                                                       \\ \hline
Trainset:Testset Size Range & \multicolumn{2}{c|}{300:100 - 1,500:500}                                        & \multicolumn{2}{c|}{3,000:1,000 - 9,000:3,000}                  \\ \hline
Neural Simulator     & \multicolumn{1}{c|}{INSIGHT} & \multicolumn{1}{c|}{FC-Ensemble in CRONuS} & \multicolumn{1}{c|}{INSIGHT} & FC-Ensemble in CRONuS \\ \hline
Two-Stage Miller-Compensated OPAMP with NMOS Input Pair    & \multicolumn{1}{c|}{0.00053}            & \multicolumn{1}{c|}{0.02054}            & \multicolumn{1}{c|}{0.00054}            &       0.01924      \\ \hline
Two-Stage Miller-Compensated OPAMP with PMOS Input Pair & \multicolumn{1}{c|}{0.00023}            & \multicolumn{1}{c|}{0.04267}            & \multicolumn{1}{c|}{0.00037}            &   0.04267          \\ \hline
Two-Stage Transimpedance Amplifier      & \multicolumn{1}{c|}{0.00040}            & \multicolumn{1}{c|}{0.04335}            & \multicolumn{1}{c|}{0.00024}            &     0.04475  \\ \hline
Three-Stage Transimpedance Amplifier    & \multicolumn{1}{c|}{0.00012}            & \multicolumn{1}{c|}{0.01688}            & \multicolumn{1}{c|}{0.00018}            &     0.01727      \\ \hline
Comparator       & \multicolumn{1}{c|}{0.00666}            & \multicolumn{1}{c|}{0.01856}            & \multicolumn{1}{c|}{0.00160}            &       0.01856      \\ \hline
Level Shifter    & \multicolumn{1}{c|}{0.00160}            & \multicolumn{1}{c|}{0.04436}            & \multicolumn{1}{c|}{0.00047}            &       0.05950      \\ \hline
\end{tabular}}
\label{tab_comparison}
\end{table*}

\begin{table*}[t]
\footnotesize
\caption{INSIGHT Performance Across Different Technologies (Training from Scratch)}
\resizebox{\textwidth}{!}{
\begin{tabular}{|c|ccc|ccc|ccc|}
\hline
Technology & \multicolumn{3}{c|}{45nm}                                            & \multicolumn{3}{c|}{130nm}                                           & \multicolumn{3}{c|}{180nm}                                           \\ \hline
Circuits  & \multicolumn{1}{c|}{Train-Test Size} & \multicolumn{1}{c|}{R\textsuperscript{2}} & MSE & \multicolumn{1}{c|}{Train-Test Size} & \multicolumn{1}{c|}{R\textsuperscript{2}} & MSE & \multicolumn{1}{c|}{Train-Test Size} & \multicolumn{1}{c|}{R\textsuperscript{2}} & MSE \\ \hline
\multirow{2}{*}{\shortstack{Two-Stage Miller-Compensated \\ OPAMP with NMOS Input Pair}}    & \multicolumn{1}{c|}{1500-500}         & \multicolumn{1}{c|}{0.9901}   &  0.00053    & \multicolumn{1}{c|}{1500-500}         & \multicolumn{1}{c|}{0.9988}   &  0.00027   & \multicolumn{1}{c|}{1500-500}         & \multicolumn{1}{c|}{0.9961}   &    0.00046 \\ \cline{2-10} 
                                         & \multicolumn{1}{c|}{3000-1000}       & \multicolumn{1}{c|}{0.9960}   & 0.00054     & \multicolumn{1}{c|}{3000-1000}       & \multicolumn{1}{c|}{0.9990}   &   0.00082 & \multicolumn{1}{c|}{3000-1000}       & \multicolumn{1}{c|}{0.9995}   &   0.00094    \\ \hline
\multirow{2}{*}{\shortstack{Two-Stage Miller-Compensated \\ OPAMP with PMOS Input Pair}}  & \multicolumn{1}{c|}{1500-500}         & \multicolumn{1}{c|}{0.9937}   &  0.00023    & \multicolumn{1}{c|}{1500-500}         & \multicolumn{1}{c|}{0.9982}   &  0.00041   & \multicolumn{1}{c|}{1500-500}         & \multicolumn{1}{c|}{0.9971}   &    0.00011     \\ \cline{2-10} 
                                         & \multicolumn{1}{c|}{3000-1000}       & \multicolumn{1}{c|}{0.9973}   & 0.00037     & \multicolumn{1}{c|}{3000-1000}       & \multicolumn{1}{c|}{0.9951}   &   0.00010 & \multicolumn{1}{c|}{3000-1000}       & \multicolumn{1}{c|}{0.9961}   &  0.00019     \\ \hline
\multirow{2}{*}{\shortstack{Two-Stage \\ Transimpedance Amplifier}}       & \multicolumn{1}{c|}{1500-500}         & \multicolumn{1}{c|}{0.9982}   &  0.00040    & \multicolumn{1}{c|}{1500-500}         & \multicolumn{1}{c|}{0.9975}   &  0.00065   & \multicolumn{1}{c|}{1500-500}         & \multicolumn{1}{c|}{0.9901}   &    0.00050     \\ \cline{2-10} 
                                         & \multicolumn{1}{c|}{3000-1000}       & \multicolumn{1}{c|}{0.9933}   & 0.00024    & \multicolumn{1}{c|}{3000-1000}       & \multicolumn{1}{c|}{0.9991}   &   0.00023 & \multicolumn{1}{c|}{3000-1000}       & \multicolumn{1}{c|}{0.9980}   &   0.00047     \\ \hline
\multirow{2}{*}{\shortstack{Three-Stage \\ Transimpedance Amplifier}}     & \multicolumn{1}{c|}{1500-500}         & \multicolumn{1}{c|}{0.9997}   &  0.00012   & \multicolumn{1}{c|}{1500-500}         & \multicolumn{1}{c|}{0.9956}   &  0.00085   & \multicolumn{1}{c|}{1500-500}         & \multicolumn{1}{c|}{0.9969}   &   0.00011   \\ \cline{2-10} 
                                         & \multicolumn{1}{c|}{3000-1000}       & \multicolumn{1}{c|}{0.9996}   & 0.00018     & \multicolumn{1}{c|}{3000-1000}       & \multicolumn{1}{c|}{0.9977}   &   0.00011 & \multicolumn{1}{c|}{3000-1000}       & \multicolumn{1}{c|}{0.9955}   &   0.00019  \\ \hline
\end{tabular}
}
\label{tab_diff_tech}
\end{table*}

\begin{table*}[t]
\footnotesize
\centering
\caption{Performance Comparison of INSIGHT-M (with PPO) vs AutoCkt for 45nm case}
\begin{tabular}{|c|c|c|c|}
\hline
Algorithm & \multicolumn{2}{c|}{INSIGHT-M (with PPO)} & AutoCkt (PPO) \\
\cline{1-4}
Circuits & Pre-training Data & Real-time Simulations & Real-time Simulations \\
\hline
Two-Stage Miller-Compensated OPAMP with NMOS Input Pair & 1,500 & \(2 \pm 0.18\) steps & 14,100 steps \\
\hline
Two-Stage Miller-Compensated OPAMP with PMOS Input Pair & 1,500 & \(2 \pm 0.18\) steps & 9,300 steps \\
\hline
Two-Stage Transimpedance Amplifier & 300 &   \(15 \pm 5\) steps & 11,100 steps \\
\hline
Three-Stage Transimpedance Amplifier & 600 &  \(15 \pm 5\) steps & 8,700 steps \\
\hline
Comparator & 1,200 &  \(15 \pm 5\) steps & 8,400 steps \\
\hline
\end{tabular}
\label{tab_RL}
\end{table*}

\subsection{Benchmark Circuits and Technologies}\label{Benchmarks}
We trained and evaluated INSIGHT on a machine with eight NVIDIA P100 GPUs and an Intel Xeon E5-2698 V4 CPU, using the following benchmark circuits to validate our approach:
2-Stage Miller-Compensated OPAMP with NMOS Input Pair,
2-Stage Miller-Compensated OPAMP with PMOS Input Pair,
2-Stage TIA (Transimpedance Amplifier),
3-Stage TIA,
Comparator, and 
Level Shifter. 
Figure~\ref{fig:circuits_bench} illustrates the detailed topology of the benchmark circuits. 
Table~\ref{tab_perf_specs} lists the performance metrics for each circuit we focussed on in this paper.
We have validated our concept using BSIM 45nm, Skywater 130nm, BSIM 180nm, and FreePDK 45nm technologies.

\subsection{INSIGHT Performance}\label{INSIGHT_Perf}
\subsubsection{Prediction Accuracy}\label{pred_acc}
Table~\ref{tab:hyperparam_fcensemble}
provides the architectural details for the FC Ensemble-based neural simulator used in CRONuS  ~\cite{DATE24:CRONus:Oh}, which serves as the baseline for our comparison with INSIGHT.
We also validate INSIGHT across various types of performance metrics.
As illustrated in Table~\ref{tab_perf_specs}, different combinations of economic metrics and simulation-intensive transient specifications have been considered for the comparator and level shifter circuits.
Table~\ref{tab_comparison} demonstrates that INSIGHT consistently achieves high R\textsuperscript{2} scores and low MSE losses across all benchmark circuits on 45nm technologies. 
The table further details INSIGHT’s performance across various training dataset sizes, indicating robust performance for dataset sizes $\sim$ 1500-1800. 
Given that simulation data collection is a one-time effort, this investment can prove cost-effective if a topology is to be optimized multiple times.
In our optimization framework, we define the initial dataset size as a hyperparameter to efficiently balance computational resource utilization with subsequent sample efficiency, allowing for adjustments tailored to specific use cases.
INSIGHT also demonstrates high prediction accuracies for simulation-intensive transient specifications given economic metric information. 
For example, using just DC Power information significantly improves INSIGHT's prediction for Avg. Delay in comparators. Similarly, using just DC Power and Ratio information significantly improves INSIGHT's prediction accuracy for Avg. Delay and Delay Balance.
Table~\ref{tab_diff_tech} presents INSIGHT's prediction accuracy scores across 45nm, 130nm, and 180nm technologies when trained from scratch, using different dataset sizes. 
By employing simple transfer learning, we could further reduce train dataset requirements.

\subsubsection{Inference Speed}\label{inf_speed}
We evaluated INSIGHT's performance prediction inference time across our benchmark circuits. When simultaneously processing a batch of 1,000 combinations of input parameters, the average inference time across circuits was $115.21\pm 12.01\mu s $.  
This result was derived from 100 trials per circuit on our setup.

\subsection{INSIGHT-M Performance}\label{INSIGHT-M_Perf}
We evaluate the performance of INSIGHT-M against AutoCkt~\cite{DATE20:Settaluri:Autockt}, an established online learning-based sizing algorithm based on PPO, using default hyperparameter settings.
Table~\ref{tab_RL} presents the performance comparison for different circuits on 45nm technology, each row representing a test case corresponding to a fixed target specification randomly selected from a set of designer-specified performance specifications similar to those used in CRONuS~\cite{DATE24:CRONus:Oh} and AutoCkt. 
Performance is measured based on the number of real-time simulations (steps) required, with fewer steps indicating better performance (sample efficiency).
The performance is evaluated over 30 trials to ensure robust statistical analysis, with INSIGHT-M's reported as mean $\pm$ standard deviation.
The table also depicts the initial dataset size used to pre-train INSIGHT for each case.
Without pre-training, the framework behaves like a pure Model-based RL setup, with the total number of steps required falling within this mentioned pre-training dataset size for each case.
As the table demonstrates, pre-trained INSIGHT enables INSIGHT-M to achieve convergence at only < 20 real-time simulations across circuits for the given RL algorithm, achieving over 100-1000X improvement in sample efficiency over existing sizing methods. 
\vspace{-10pt}
\section{Conclusion} \label{sec:conclusion}
This paper introduces INSIGHT, a novel, GPU-powered, technology-agnostic, effective universal neural simulator as a low-cost and high-fidelity substitute for standard circuit simulators in the analog design automation loop.
INSIGHT facilitates \textit{effective} and \textit{efficient} design space exploration for improved sample efficiency.
Leveraging self-attention to understand underlying inherent parameter-performance interdependencies within a given topology, INSIGHT accurately predicts performance at microsecond inference times across diverse circuit types and technologies.
INSIGHT also enables accurate prediction of simulation-costly critical transient specifications using less expensive performance metric information. 
Our Model-Based Batch RL framework, INSIGHT-M, which utilizes INSIGHT for analog sizing, only requires < 20 real-time simulations with 100-1000x lower simulation costs across circuits, showcasing the effectiveness of INSIGHT.
To the best of our knowledge, this marks the first successful application of a universal neural simulator framework leveraging autoregressive Transformers for analog design automation.
We also plan to extend this technology to address more complex analog design challenges, including mismatch analysis and technology design migration.


{
\bibliographystyle{unsrt}

\bibliography{ref/Top_sim, ref/reference, ref/ref}

}

\end{document}